\documentclass[final]{cvpr}
\usepackage{booktabs}
\usepackage{times}
\usepackage{epsfig}
\usepackage{graphicx}
\usepackage{amsmath}
\usepackage{amssymb}
\usepackage{xcolor}
\usepackage{caption}
\usepackage{subcaption}
\usepackage{comment}
\usepackage[normalem]{ulem}

\usepackage{hyperref}
\hypersetup{pagebackref=true,breaklinks=true,colorlinks,bookmarks=false}

\newcommand{\methodname}{CADDY}
\def\cvprPaperID{8433} 
\def\confYear{CVPR 2021}

\begin{document}

\title{Playable Video Generation}

\author{Willi Menapace\\
\normalsize University of Trento \\
{\tt\small willi.menapace@unitn.it}
\and
St\'{e}phane Lathuili\`{e}re\thanks{The second and third authors contributed equally to the work.}\\
\normalsize LTCI, T\'{e}l\'{e}com Paris\\
\normalsize Institut Polytechnique de Paris \\
{\tt\small stephane.lathuiliere@telecom-paris.fr}
\and
Sergey Tulyakov\footnotemark[1]\\
\normalsize Snap Inc. \\
{\tt\small stulyakov@snap.com}

\and
Aliaksandr Siarohin\\
\normalsize University of Trento \\
{\tt\small aliaksandr.siarohin@unitn.it}

\and
Elisa Ricci\\
\normalsize University of Trento \\
\normalsize Fondazione Bruno Kessler \\
{\tt\small e.ricci@unitn.it}
\vspace{-0.7cm}
}


\maketitle

\begin{abstract}
  This paper introduces the unsupervised learning problem of playable video generation (PVG).
  In PVG, we aim at allowing a user to control the generated video by selecting a discrete action at every time step as when playing a video game.
  The difficulty of the task lies both in learning semantically consistent actions and in generating realistic videos conditioned on the user input. 
  We propose a novel framework for PVG that is trained in a self-supervised manner on a large dataset of unlabelled videos. We employ an encoder-decoder architecture where the predicted action labels act as bottleneck. The network is constrained to learn a rich action space using, as main driving loss, a reconstruction loss on the generated video. We demonstrate the effectiveness of the proposed approach on several datasets with wide environment variety. Further details, code and examples are available on our project page \href{https://willi-menapace.github.io/playable-video-generation-website/}{willi-menapace.github.io/playable-video-generation-website}.   
  
\end{abstract}

\vspace{-3mm}
\section{Introduction}

Humans at a very early age can identify key objects and how each object can interact with its environment. This ability is particularly notable when watching videos of sports or games. In tennis and football, for example, the skill is taken to the extreme. Spectators and sportscasters often argue which action or movement the player should have performed in the field.
We can understand and anticipate actions in videos despite never being given an explicit list of plausible actions. We develop this skill in an unsupervised manner as we see actions live and on the screen. We can further analyze the technique with which an action is performed as well as the “amount” of action, i.e. how much to the left. Furthermore, we can reason about what happens if the player took a different action and how this would change the video.

\setlength{\belowcaptionskip}{0pt}
\begin{figure}
     \centering
     \begin{subfigure}[b]{0.58\columnwidth}
         \centering
         \includegraphics[width=\textwidth]{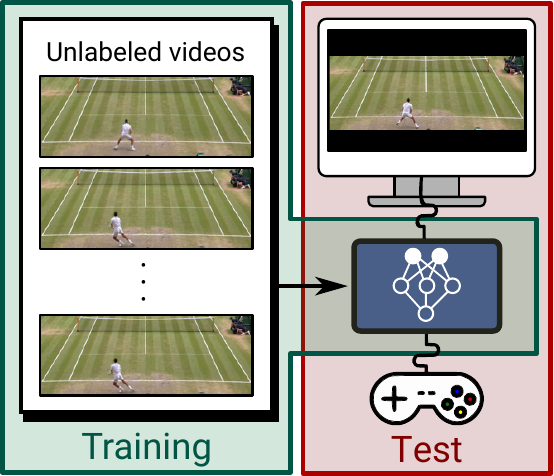}
         \caption{Playable Video Generation}
         \label{fig:teaser_a}
     \end{subfigure}
     \hfill
     \begin{subfigure}[b]{0.32\columnwidth}
         \centering
         \includegraphics[width=\textwidth]{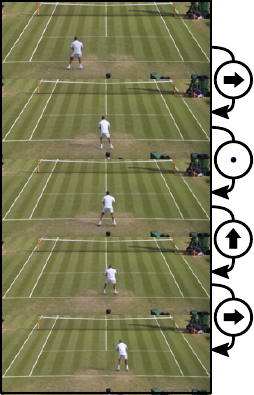}
         \caption{Our results}
         \label{fig:teaser_b}
     \end{subfigure}
     \caption{We introduce the task of playable video generation in an unsupervised setting (left). Given a set of unlabeled video sequences, a set of discrete actions are learned in order to condition video generation. At test time, using our method, named \methodname, the user can control the generated video on-the-fly providing action labels.}
\label{fig:teaser}
        \vspace{-0.5cm}
        \end{figure}

\setlength{\belowcaptionskip}{-10pt}


From this observation, we propose a new task, Playable Video Generation (PVG) illustrated in Fig~\ref{fig:teaser_a}. In PVG, the goal is to learn a set of distinct actions from real-world video clips in an unsupervised manner (green block) in order to offer the user the possibility to interactively generate new videos (red block). As shown in Fig~\ref{fig:teaser_b}, at test time, the user provides a discrete action label at every time step and can see live its impact on the generated video, similarly to video games. Introducing this novel problem paves the way toward methods that can automatically simulate real-world environments and provide a gaming-like experience.  

PVG is related to the future frame prediction problem~\cite{franceschi2020stochastic,lee2018savp,mathieu2015deep,srivastava2014dropout,tulyakov2018moco,vondrick2015anticipating} and in particular to methods that condition future frames on action labels~\cite{kim2020_gamegan,nunes2020action,oh2015action}. 
Given one or few initial frames and the labels of the performed actions, such systems aim at predicting what happens next in the video. For example, this framework can be used to imitate a desired video game using a neural network with a remarkable generation quality~\cite{kim2020_gamegan,oh2015action}. 
However, at training time, these methods require videos with their corresponding frame-level action at every time step. Consequently, these methods are limited to video game environments~\cite{kim2020_gamegan,oh2015action} or robotic data~\cite{nunes2020action} and cannot be employed in real-world environments. 
As an alternative, the annotation effort required to address real-world videos can be reduced using a single action label to condition the whole video \cite{WANG_2020_WACV}, but it limits interactivity since the user cannot control the generated video on-the-fly. Conversely, in PVG, the user can control the generation process by providing an action at every time-step after observing the last generated frame.



This paper addresses these limitations introducing a novel framework for PVG named \emph{Clustering for Action Decomposition and DiscoverY} (CADDY). Our approach discovers a set of distinct actions after watching multiple videos through a clustering procedure blended with the generation process that, at inference time, outputs playable videos.
We adopt an encoder-decoder architecture where a discrete bottleneck layer is employed to obtain a discrete representation of the transitions between consecutive frames. A reconstruction loss is used as main driving loss avoiding the need for neither action label supervision nor even the precise number of actions.
A major difficulty in PVG is that discrete action labels cannot capture the stochasticity typical of real-world videos. To address this difficulty, we introduce an action network that estimates the action label posterior distribution by decomposing actions in a discrete label and a continuous component. While the discrete action label captures the main semantics of the performed action, the continuous component captures how the action is performed. At test time, the user provides only the discrete actions to interact with the generation process.

Finally, we experimentally demonstrate that our approach can learn consistent actions in varied environments. We conduct experiments on three different datasets including both real-world (\ie tennis and robotics \cite{ebert2017selfsupervised}) and synthetic (\ie video game \cite{brockman2016gym}). Our experiments show that \methodname~generates high-quality videos while offering the user a better playability experience than existing future frame prediction methods.




\section{Related Works}
\noindent\textbf{Video generation.} Recent advances in deep generative models have led to impressive progress in video generation.
A variety of formulations have been explored including Generative Adversarial Networks (GANs)~\cite{saitotrain,tulyakov2018moco,vondrick2015anticipating}, variational auto-encoders (VAEs)~\cite{babaeizadeh2018stochastic} and auto-regressive models~\cite{weissenborn2020scaling}.
While early works addressed the unconditional case \cite{saitotrain,vondrick2015anticipating}, many conditional video generation tasks have been addressed.
The generated video can be conditioned on specific type of information.
Among numerous examples, we can cite video to video translation~\cite{wang2018video}, image animation~\cite{Siarohin2019firstorder} or pose-based generation \cite{chan2019everybody}.


Another typical example is the problem of future frame prediction that consists in generating a video conditioned on the first video frames. Early works employ deterministic predictive models \cite{finn2016cdna,mathieu2015deep,vondrick2015anticipating} that cannot handle the stochasticity inherent to real-world videos. Several methods have been proposed to incorporate stochasticity in the model using VAE formulations~\cite{franceschi2020stochastic,lee2018savp,tulyakov2018moco}, GANs~\cite{kwon2019predicting}, or normalizing flows~\cite{kumar2019videoflow}. Inspired by VAE-based approaches, we adopt a probabilistic formulation that is compatible with our encoder-decoder pipeline. Nevertheless, our playable video generation framework goes beyond video prediction methods and allows the user to control generation of future frames with discrete actions.

To offer better control over the generation process than video prediction methods, some approaches condition the generated video on an input action label \cite{kim2019unsupervisedkeypointlearning,WANG_2020_WACV}. However, these approaches require action annotations for training and the generated video cannot be controlled on-the-fly since a single label is used to generate the whole video. To further increase the level of control, other approaches condition each frame in the generated video on a different action label \cite{chiappa2017recurrent,kim2020_gamegan,nunes2020action,oh2015action}. In particular, Kim \etal recently proposed \emph{GameGAN}~\cite{kim2020_gamegan}, a framework that  visually imitates a desired game. During training, GameGAN ingests a large collection of videos and the corresponding keyboard actions pressed by the player. The network is trained to predict the next frame from past frames and keyboard actions. While these methods allow playable video generation similarly to our approach, the requirement for frame-level annotation limits their application to constrained environments such as video games. Conversely, our approach does not use action label supervision and can be employed in real-world environments. In the context of future frame predictions, Rybkin \etal~\cite{rybkin2018learning} propose to infer latent actions. However, this method learns a continuous action space that is later mapped to discrete actions with action supervision.

Few works have focused on the generation of videos with controllable characters in real-world environments \cite{gafni2020vid2game,zhang2020vid2player}. 
 However, these approaches heavily rely on prior knowledge specific to the environment, such as pre-trained full-body pose estimators. 
On the contrary, in PVG, the goal is to design a general framework that can be applied to varied environments without modifications.



 \noindent\textbf{Deep clustering and unsupervised learning.} Unsupervised learning has attracted growing attention with the raise of self-supervised learning techniques. 
 Recent works address jointly clustering and representation learning so that the network is trained to categorize the training data while learning deep representation.
 For instance, DeepCluster~\cite{DBLP:journals/corr/abs-1807-05520} alternates between k-mean clustering and a feature learning phase where cluster assignments are used as supervision.
 As an alternative to this iterative algorithm, Ji \etal propose Invariant Information Clustering (IIC)~\cite{DBLP:journals/corr/abs-1807-06653} where the networks directly outputs cluster assignment rather than  feature representations. In IIC, the network is trained with an information-theoretic criterion formulated as a mutual information maximization problem. Differently from these approaches that focus only on image clustering, CADDY tackles the problem of learning actions in videos by blending clustering with a generation formulation. 

 Note that, recent works address the problem of unsupervised learning for videos \cite{dwibedi2019temporal,luo2017unsupervised,peng2020unsupervised,soomro2017unsupervised} but they are either limited to representation learning ~\cite{dwibedi2019temporal,luo2017unsupervised} or perform clustering at the video-level \cite{peng2020unsupervised,soomro2017unsupervised} while PVG requires action labels at every time step. 
 
\begin{figure*}[ht]
    \centering
    \includegraphics[width=0.85\textwidth]{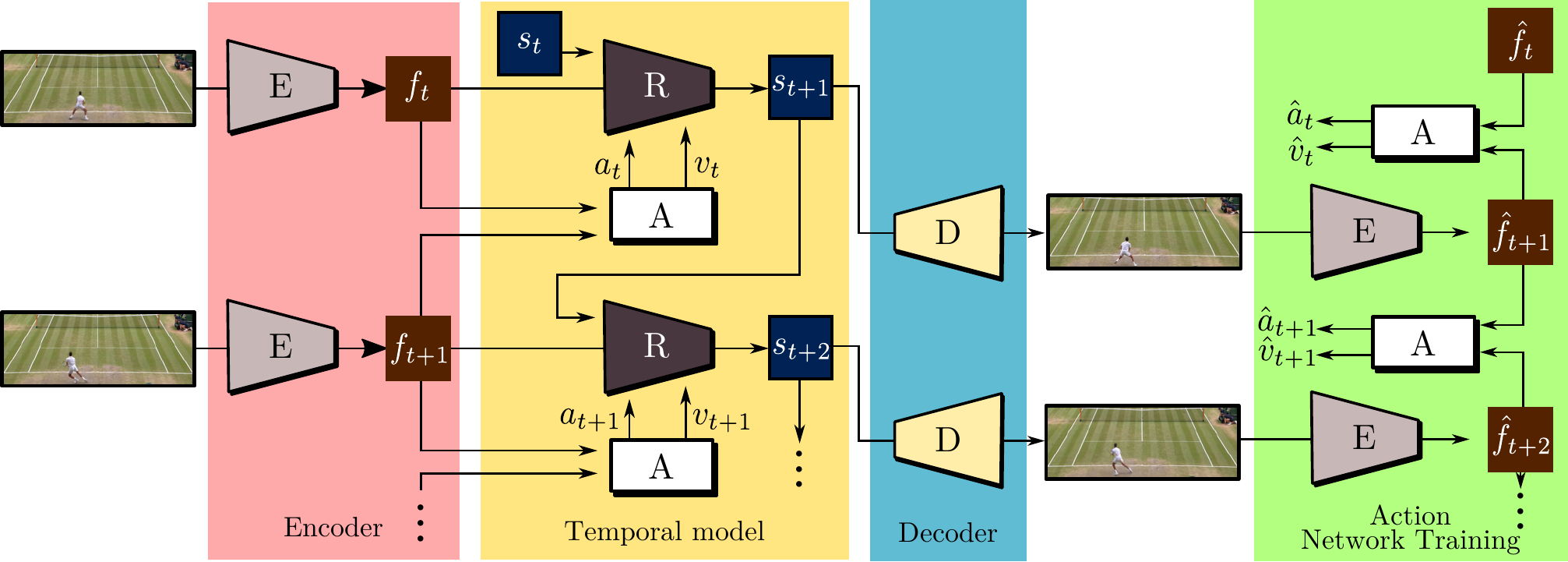}
    \caption{\methodname's training procedure for unsupervised playable video generation. An encoder $E$ extracts frame representations from the input sequence. A temporal model estimates the successive states using a recurrent dynamics network $R$ and an action network $A$ which predicts the action label corresponding to the current action performed in the input sequence. Finally, a decoder $D$ reconstructs the input frames. The model is trained using reconstruction as the main driving loss.}
    \label{fig:pipeline}
\end{figure*}

\setlength{\abovedisplayskip}{3pt}
\setlength{\belowdisplayskip}{3pt}

\section{Method}
\subsection{Overall Pipeline}
In this work, we propose a framework where a user can interactively control the video generation process by selecting an action at every time step among a set of $K$ discrete actions. Our method, named CADDY, is trained on a dataset of unannotated videos. We only assume that the video sequences depict a single agent acting in an environment. No action labels are required. 

Inspired by Reinforcement Learning (RL) literature ~\cite{sutton2018reinforcement}, the object in the scene is modeled as an agent interacting with its environment by performing an action at every time step. Differently from RL, our goal is to jointly learn the action space, the state representation that describes the state of the agent and its environment, and a decoder that reconstructs the observations (\ie frames) from the state.


CADDY is articulated into four main modules illustrated in Fig.~\ref{fig:pipeline}: (i) an encoder employs a network $E$ to extract frame representations; (ii) a temporal model estimates the label corresponding to the action performed in the current frame and predicts a state $s_{t+1}$ that describes the environment at the next time step, after performing the detected action. The action label is predicted via a network $A$ that receives the frame representations from the current and next frames. To predict the next frame environment state $s_{t+1}$, we employ a recurrent neural network $R$ that we refer to as \emph{Dynamics network}. (iii) a decoder module employs a network $D$ to reconstruct each frame from the frame embedding predicted by the temporal model. (iv) the reconstructed frames are fed to the encoder to assess the quality of the estimated action labels. 

The overall pipeline is trained in an end-to-end fashion using as main driving loss a reconstruction loss on the output frames. The key idea of our approach is that the action network $A$ needs to predict consistent action labels in order to correctly estimate the next frame embeddings $s_{t+1}$, and then, accurately reconstruct the input frames. We now describe each network.



\vspace{1.0mm}
\noindent\textbf{Video encoder.} Let us consider a video sequence $\{x_t\}_{t=1}^T$ of length $T$. First, for every frame, we extract a feature representation $f_t$ with $t\in\{1,..,T\}$ which we call \textit{frame features}. The aim of this representation is to encode information about the appearance and the semantics of the input frame:
\vspace{-1.5mm}
\begin{equation}
f_t=E(x_t).
\end{equation}

\vspace{-1.4mm}

\noindent\textbf{Action network.} Next, the action network $A$ is used to infer the discrete \textit{action} $a_t\in \{1..K\}$ performed between the frames $x_t$ and $x_{t+1}$.  The action label $a_t$ characterizes the transition between the features $f_t$ and its successor $f_{t+1}$. In a deterministic environment, the evolution of the environment is fully described by $a_t$. However, in real scenarios this assumption is rarely satisfied and agents acting in a complex environment may express a behavior that can only be partially described with discrete actions. To handle this, we propose to model the transition between two consecutive frame features as the combination of the discrete action $a_t$ and a continuous component $v_t$ referred to as \emph{action variability embedding}, \textit{i.e.}:
\begin{equation}
a_t, v_t = A(f_t,f_{t+1}).
\end{equation}
While the discrete action $a_t$ describes the action performed by the agent at a high level, the continuous component $v_t$ describes the variability of each action and captures the possible non-determinism in the environment.


In order to handle this non-determinism, we propose a probabilistic auto-encoder formulation to train the action network. In this regard, details are given in Sec.~\ref{sec:action_network}.




\vspace{1.0mm}

\noindent\textbf{Dynamics network.} Given the features $f_t$, the action $a_t$, and the corresponding \emph{action variability embedding} $v_t$, the dynamics network $R$ is used to embed the dynamics of the environment observed in the input sequence. We model the dynamics network as a recurrent network based on convolutional LSTMs \cite{shi2015convlstm} with the \textit{environment state} $s_t$:
\vspace{0.5mm}
\begin{equation}
    s_{t+1} = R(s_t, f_t, a_t, v_t).\label{eq:rec}
\end{equation}
\vspace{-2.9mm}



\noindent The induction in Eq. \eqref{eq:rec} is initialized with a parameter $s_0$ that we consider as a trainable parameter. The action label $a_t$ and the \emph{action variability embedding} $v_t$ are concatenated channel-wise with the input feature $f_t$ in the dynamics network. 

\vspace{1.0mm}
\noindent\textbf{Decoder.} We use a decoder network $D$ to reconstruct the frames $\hat{x}_{t+1}$ from the \emph{environment state} $s_{t+1}$.


\subsection{Probabilistic Action Network}
\label{sec:action_network}

As introduced earlier, the goal of the action network is to estimate the action label $a_t$ and \emph{action variability embedding} $v_t$. 
The action network first uses an \emph{action state network} $A_s$ to extract an \emph{action embedding} $e_t$ which represents the information in $f_t$ that is related to the action currently performed. 
Following a probabilistic formulation, $A_s$ predicts the posterior distribution of the \emph{action embedding} modeled as Gaussian:
\begin{align}
&e_t \sim \mathcal{N}(\mu_{e_t}, \sigma_{e_t}^2)\\
\text{with~} &\mu_{e_t}, \sigma_{e_t}^2 = A_s(f_t).
\end{align}
Similarly, the distribution of the \emph{action embedding} $e_{t+1}{\sim} \mathcal{N}(\mu_{e_{t+1}}, \sigma_{e_{t+1}}^2)$ is estimated from $f_{t+1}$. 
We can then combine the action states $e_{t}$ and $e_{t+1}$ to predict the performed action. More precisely, we propose to model the posterior distribution of the action $(a_t,v_t)$ using the difference between the two \emph{action embeddings} which still follows a Gaussian distribution. From the independence between random variables $e_t$ and $e_{t+1}$ we have: 
\begin{align}
&d_t = e_{t+1} - e_{t} \sim \mathcal{N}(\mu_{d_t}, \sigma_{d_t}^2)\label{eq:gau}\\
\text{with } &\mu_{d_t}=\mu_{e_{t+1}} - \mu_{e_t} \text{~and~} \sigma_{d_t}^2=\sigma_{e_{t+1}}^2 {+}\sigma_{e_t}^2.
\end{align}
Finally, the \emph{action direction} $d_t$ is sampled according to \eqref{eq:gau} using the re-parametrization trick~\cite{kingma2014auto} and fed to a single-layer classifier to estimate the probability of each action $p_t{=}\{p_t^k\}_{k=1}^K{\in} [0,1]^K$. Importantly, we implement the classification layer using a Gumbel-Softmax layer \cite{jang2017gumbel} to obtain the discrete label $a_t$ from the probabilities $p_t$ while preserving differentiability for Stochastic Gradient Descent (SGD).


Regarding the \emph{action variability embedding} $v_t$, considering $v_t{=} d_t$ would not favour the model to learn meaningful action labels $a_t$ since the changes in the environment could be directly encoded in $d_t$ without using $a_t$. Thus, we propose to make $v_t$ dependent on both $d_t$ and $a_t$ in a way that $d_t$ cannot be recovered from $v_t$ alone, enforcing the learning of action labels $a_t$. 
To this aim, we consider a set of $K$ \textit{action direction centroids} $\{c_k\}_{k=1}^{K}$ (one per action) that are defined as the expected \emph{action directions} for each action. Practically, the cluster centroids are estimated using an exponential moving average over the action direction associated with each discrete action. 
We propose to define the \emph{action variability embedding} $v_t$ as the difference between the observed action direction $d_t$ and its assigned cluster centroid.
Following a soft-assignement formulation, the \emph{action variability embedding} is given by the expected difference with its action centroid:
\vspace{-1.5mm}
\begin{equation}
v_t = \sum_{k=1}^{K} p_t^k(d_t - c_k).
\end{equation}
In this way, $v_t$ acts as a bottleneck for $d_t$ since, assuming distinct centroids $c_k$, $d_t$ cannot be completely encoded in $v_t$. To ensure distinct centroids, we introduce a loss that prevents their collapse to a single point (see Sec.~\ref{sec:loss_ac}).


\subsection{Training objectives and test}

CADDY is trained in an end-to-end fashion with a combination of objectives that aim at obtaining both high-quality output sequences and a discrete action space that captures the agent's high-level actions.

\noindent\textbf{Reconstruction losses.}
The main driving loss of our system is a frame reconstruction loss based on the perceptual loss of Johnson \etal{} \cite{johnson2016perceptual}:
\begin{equation}
\mathcal{L}_\mathrm{rec}^{x} = \frac{1}{T} \sum_{t=1}^{T}\sum_{j=1}^{J}\left\|N_j(x_t) - N_j(\hat{x}_t)\right\|_1,
\end{equation}
where $N_j$, $j \in \{1, ..., J\}$ indicates the $j^{th}$ channel extracted from a specific VGG-19 layer and $J$ is the number of
feature channels in this layer. 
The reconstruction loss is averaged over different layers and resolutions~\cite{Siarohin2019firstorder}: $D$ is equipped with multiple heads to output $\hat{x}_t$ at different resolutions and $x_t$ is down-sampled to form a pyramid. This loss is completed by a $L_1$ reconstruction loss (denoted as $\mathcal{L}_1$) between $\hat{x}_t$ and $x_t$ to improve convergence.  


Additionally, we propose to assess reconstruction quality at the {frame feature} and action levels. The computation of these additional losses is represented in the green block of Fig.~\ref{fig:pipeline}. The reconstructed frame $\hat{x_t}$ is fed to $E$ leading to {frame features} $\hat{f}_t$. Then, the Euclidean distance between initial and reconstructed features is employed:
\begin{align}
\mathcal{L}_\mathrm{rec}^f = \frac{1}{T} \sum_{t=1}^{T}\left\|f_t - \hat{f}_t\right\|_2^2. \label{eq:recf}
\end{align}
The motivation for this loss is that it enforces that the semantic information extracted from input frames $x_t$ is also preserved in $\hat{x}_t$. 
To avoid trivial minimization of \eqref{eq:recf}, we do not back-propagate the loss gradient through $f_t$.

\noindent\textbf{Action losses.}
\label{sec:loss_ac}
\noindent Regarding action understanding, we propose an information-theoretic objective that acts on the action probabilities. The reconstructed \emph{frame features} $\hat{f}_t$ and $\hat{f}_{t+1}$ are fed to $A$ that internally estimates the action probabilities $\hat{p}_t$. 
We then  maximize the mutual information between the actions extracted from the input sequences and the corresponding ones extracted from the reconstructed sequences:
\begin{align}
  \max_{\theta}\mathcal{MI}(p_t,\hat{p}_t) = \max_{\theta}(\mathcal{H}(p_t) - \mathcal{H}(p_t|\hat{p}_t))\label{eq:MI}.
\end{align}
Maximizing this objective function has two desirable effects. First, minimizing the conditional entropy term $\mathcal{H}(p_t|\hat{p}_t)$ imposes that the same action must be inferred from the input and the reconstructed sequence. Second, maximizing the entropy term $\mathcal{H}(p_t)$ enforces that the maximal number of actions must be discovered. This avoids trivial solutions where only a single action label $a_t$ is constantly predicted and ensures that the $K$ \emph{action direction centroids} $\{c_k\}_{k=1}^K$ do not collapse to a single point.

To compute the mutual information in Eq.~\eqref{eq:MI}, we consider a mini-batch $\mathcal{B}$ of action probabilities ($p_t$,$\hat{p}_t$). We estimate the joint probability matrix $P_{ij}{=}P(a_t{=}i, \hat{a}_t{=}j)$, with $(i,j) \in \{1,..., K\}^2$, as follows:  
\begin{align}
  P_{ij}=P(a_t{=}i, \hat{a}_t{=}j)=\frac{1}{\bar{\mathcal{B}}}\sum_{(p_t,\hat{p}_t)\in\mathcal{B}} p_t.{\hat{p}_t}^\top,
\end{align}
where $\bar{\mathcal{B}}$ denotes the size of $\mathcal{B}$. We obtain $P_{i}{=}P(a_t{=}i)$ and $P_{j}{=}P(\hat{a}_t{=}j)$ by marginalization over $P_{ij}$. The mutual information loss term is then given by:
\begin{align}
  \mathcal{L}_\mathrm{act} = -\mathcal{MI}(p_t,\hat{p}_t)=\sum_{t=1}^{K}\sum_{j=1}^{K} P_{ij} \ln\frac{P_{ij}}{P_{i}.P_{j}}.
\end{align}
This action matching loss $\mathcal{L}_\mathrm{act}$  is completed by a loss that enforces that the \emph{action variability embeddings} match. Considering the \emph{action direction} distributions $d_t{\sim}\mathcal{N}(\mu_{d_t}\sigma_{d_t}^2)$ and $\hat{d_t}{\sim}\mathcal{N}(\mu_{\hat{d}_t}, \sigma_{\hat{d}_t}^2)$ associated with each $(p_t,\hat{p}_t)$ in $\mathcal{B}$, we measure the average Kullback–Leibler (\emph{KL}) divergence between the distribution of $\hat{d}_t$ and $d_t$: 
\begin{equation}
\mathcal{L}_\mathrm{rec}^{a}{=}\frac{1}{\bar{\mathcal{B}}}\sum_{(\hat{d}_t,d_t)\in\mathcal{B}} \mathcal{D}_\mathrm{KL}(\mathcal{N}(\mu_{\hat{d}_t}, \sigma_{\hat{d}_t}^2) \| \mathcal{N}(\mu_{d_t}, \sigma_{d_t}^2)).
\end{equation}
Assuming a Gaussian prior for $d_t$, we minimize the average \emph{KL} divergence $\mathcal{D}_\mathrm{KL}$ between the distributions of $d_t$ and a Gaussian prior with unit variance $\mathcal{N}(0, I_n)$,
\begin{equation}
\mathcal{L}_\mathrm{KL} = \frac{1}{\bar{\mathcal{B}}}\sum_{d_t\in\mathcal{B}} \mathcal{D}_\mathrm{KL}(\mathcal{N}(\mu_{d_t}, \sigma_{d_t}^2) \| \mathcal{N}(0, I_n)).
\end{equation}

\noindent\textbf{Total loss.} The total objective is given by: 
\begin{align}
\mathcal{L} = &\mathcal{L}_{1} + \mathcal{L}_\mathrm{rec}^{x}+ \lambda_\mathrm{rec}^f\mathcal{L}_\mathrm{rec}^f +  \lambda_\mathrm{rec}^{a}\mathcal{L}_\mathrm{rec}^{a}\\
&+ \lambda_\mathrm{act}\mathcal{L}_\mathrm{act} +  \lambda_\mathrm{KL}\mathcal{L}_\mathrm{KL} \notag,
\end{align}
where $\lambda_\mathrm{act}, \lambda_\mathrm{rep},\lambda_\mathrm{KL}, \lambda_\mathrm{rec}^{a}$ are positive weighting parameters. These parameters are estimated on the training set: for every term, we iteratively experiment raising parameter values until we can observe that the first back-propagation steps lead to a decrease of the loss term.

\noindent\textbf{Test time.}
At test time, \methodname~ receives as input the initial video frame and the user provides an action label at every time-step. We follow an auto-regressive approach where the estimated $\hat{x}_{t+1}$ is used as input at the next time step (instead of $x_{t+1}$). We no longer employ the action network $A$ and use the action labels $a_t$ provided by the user. Regarding the \emph{action variability embedding}, we employ $v_t{=}0$ that corresponds to the maximum of its posterior distribution.

Note that, this difference between training and test would affect the performance of the network that has never received generated images as inputs. To mitigate this problem, we propose a mixed training procedure. For the first $T_f$ frames, we use the training procedure as described above where the \emph{Dynamic Network} $R$ receives in input the \emph{frame features} $f_t$ computed from original frames. Then, for $t{>}T_f$, $R$ is fed with the \emph{frame features} $\hat{f}_t$ computed from the reconstructed frames $\hat{x}_t$. By mixing original and reconstructed frames, we obtain a network that can handle both real and reconstructed frames, and consequently, is less prone to the shift issue typical of auto-regressive methods.

\section{Experiments}

\noindent\textbf{Datasets.} We evaluate our method on three video datasets:\\
\noindent \textbullet~\textit{BAIR} robot pushing dataset \cite{ebert2017selfsupervised}. We employ a version of the dataset in 256x256 resolution, composed of about 44K videos of 30 frames. 
Ground-truth information about the robotic arm position is available but is used only for evaluation purposes.\\
\noindent \textbullet~\textit{Atari Breakout} dataset. We collect a dataset  using a Rainbow DQN agent \cite{hessel2018rainbow} trained on the Atari Breakout video game environment. We collect 1407 sequences of about 32 frames with resolution 160x210  (358 for training, 546 sequences for validation and 503 for testing).\\
\noindent \textbullet~\textit{Tennis} dataset. We collect Youtube videos corresponding to two tennis matches from which we extract about 900 videos with resolution 256x96. (See \emph{Sup. Mat.} for more details).  To respect the single agent assumption, we consider only the lower part of the field.

\noindent\textbf{Evaluation Protocol.}
We propose to evaluate both action and video generation qualities by comparing the models on the video reconstruction task. A test sequence is considered and the action network is used to extract the sequence of learned discrete actions characterizing the input sequence. Starting from the initial frame, the extracted actions are used to reconstruct the remaining of the sequence. 
The evaluation is completed with a user-study that directly assesses the quality of the learned set of discrete actions.
 We adopt a large set of metrics.
 
\noindent\textbf{Video quality metrics.} \\
\noindent \textbullet~\textit{LPIPS}~\cite{zhang2018unreasonable}: We report the average \textit{LPIPS} computed on corresponding frames of input and reconstructed sequences.\\
\noindent \textbullet~\textit{FID}~\cite{heusel2017advances}: We report the average \textit{FID} between the original and reconstructed frames.\\
\noindent \textbullet~\textit{FVD}~\cite{unterthiner2018towards}: We compute the \textit{FVD} between the original and reconstructed videos.

\noindent\textbf{Action-space metrics.}
We introduce two metrics that measure the quality of the action space. Both metrics use additional knowledge (\ie ground-truth information or an externally trained detector) to measure motion consistency among frames where the same action is performed. Assuming two consecutive frames, we measure the displacement $\Delta$ of a reference point on the object of interest. On the \textit{Tennis} dataset, we employ  FasterRCNN \cite{ren2015faster} to detect the player and use the bounding box center as the reference point. On \textit{Atari Breakout}, we employ a simple pixel matching search to detect the rectangular platform and use its center as the reference point. Finally, on \emph{BAIR}, we use the ground-truth location information of the robotic arm. 
The key idea of the two action quality metrics is to assess whether the predicted action labels and the displacement $\Delta$ are consistent by trying to predict one from the other:
\\
\noindent \textbullet~\textit{$\Delta$ Mean Squared Error ($\Delta$-MSE)}: This metric measures how $\Delta$ can be regressed from the action label. For each action, we estimate the average displacement $\Delta$, which is the optimal estimator for $\Delta$ (in terms of MSE). We report the MSE to evaluate regression quality. To facilitate comparison among datasets, we normalize the MSE by the variance of $\Delta$ over the dataset.\\
\noindent \textbullet~\textit{$\Delta$-based Action Accuracy ($\Delta$-Acc)}: This metric measures how the predicted action can be predicted from the displacement $\Delta$. To this aim, we train a linear classifier and report the action-accuracy measured on the test set.

\noindent\textbf{Action-conditioning metrics.}
Finally, we consider two metrics that measure how the action label conditions the generated video. Therefore, it evaluates both generation quality and the learned action space:\\
\noindent \textbullet~\textit{Average Detection Distance (ADD)}. Similarly to \cite{Siarohin2019firstorder},  we report the Euclidean distance between the reference object keypoints in the original and reconstructed frames. We employ the same detector as for the action-quality metrics to estimate the reference points.\\ 
\noindent \textbullet~\textit{Missing Detection Rate (MDR)}: We report the percentage of detections that are successful in the input sequence, but not in the reconstructed one. This represents the proportion of frames where the object of interest is missing.

\subsection{Experimental analysis of \methodname}

\noindent\textbf{Ablation study.} In this section, we study the impact of three key elements of the proposed approach: Gumbel-Softmax sampling, the action loss $\mathcal{L}_\mathrm{act}$ and \emph{action variability embeddings} $v_t$. We produce the following variants of our method: (i) uses none of these components, (ii) introduces G.S., (iii) employs G.S. and $\mathcal{L}_\mathrm{act}$, (iv) uses G.S. and $v_t$. The \textit{BAIR} dataset is used because ground-truth displacement $\Delta$ are available.


The results are presented in Tab.~\ref{table:ablation}. 
In (i), when no component is used, the lack of G.S. sampling makes the model exploit $p_t$ at training time to encode information about the next frame, learning a continuous rather than a discrete action representation. At test time, when discrete actions are used, this mismatch leads to poor performance. When G.S. sampling is introduced in (ii), the model learns discrete action representations with a $\Delta$-\emph{MSE} of 64.8\%. However, the model lacks a mechanism to resolve the ambiguity of which variation of the discrete action should be performed. This results in difficulties in optimizing the reconstruction objective and leads to reduced video quality. In (iii), when both G.S. sampling and $\mathcal{L}_\mathrm{act}$ are present, the optimization process favors $\mathcal{L}_\mathrm{act}$ rather than the reconstruction objective, resulting in degraded performance. Lastly, in (iv), when both G.S. sampling and $v_t$ are used without $\mathcal{L}_\mathrm{act}$, the model uses $v_t$ at training time to encode complete information about the next frame and a discrete action space is not learned. At test time, when $v_t{=}0$, performance of the model is poor. Overall, this ablation study confirms the positive impact of G.S sampling, the \emph{action variability embeddings}, and our mutual-information loss on the performance.


\begin{table}
\begin{center}
\setlength{\tabcolsep}{1.8pt}
\footnotesize
\begin{tabular}{llllccccc}
\midrule
Variant &G.S &$v_t$ &$\mathcal{L}_\mathrm{act}$& LPIPS$\downarrow$ & FID$\downarrow$ & FVD$\downarrow$  & $\Delta$-\emph{MSE}$\downarrow$ & $\Delta$-\emph{Acc}$\uparrow$\\
\midrule
(i) & &&&0.263 &  80.0 & 1300 &69.7 & 51.2 \\
(ii) &\checkmark&&& 0.209  & 42.3 & 571 & 64.8 & 37.9 \\
(iii) &\checkmark&&\checkmark& 0.249  & 76.4 &  1130 & 92.7 & 24.1 \\
(iv) &\checkmark&\checkmark& & 0.245  & 76.9 & 1130 & 93.7 & 27.6 \\
\midrule

\methodname~(Full)&\checkmark&\checkmark&\checkmark&  \textbf{0.202} & \textbf{35.9} & \textbf{423} & \textbf{54.8} & \textbf{69.0} \\
\midrule
\end{tabular}
\end{center}
\caption{Ablation results on the \textit{BAIR} dataset. G.S: use of Gumbel-Softmax, $v_t$; use of the \emph{action variability embedding} $v_t$;  $\mathcal{L}_\mathrm{act}$: training with the mutual information loss $\mathcal{L}_\mathrm{act}$. $\Delta$-\emph{MSE} and $\Delta$-\emph{Acc} in \%.}
\vspace{-2mm}
\label{table:ablation}
\end{table}

\begin{figure*}[t]
     \centering
     \begin{subfigure}[b]{0.3\textwidth}
         \centering
         \includegraphics[height=2cm]{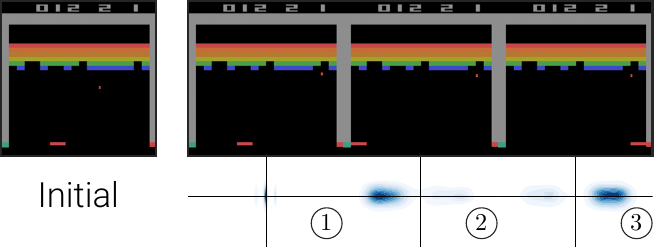}\vspace{-0.2cm}
         \caption{Atari Breakout}
     \end{subfigure}
     \hfill
     \begin{subfigure}[b]{0.63\textwidth}
         \centering
         \includegraphics[height=2cm]{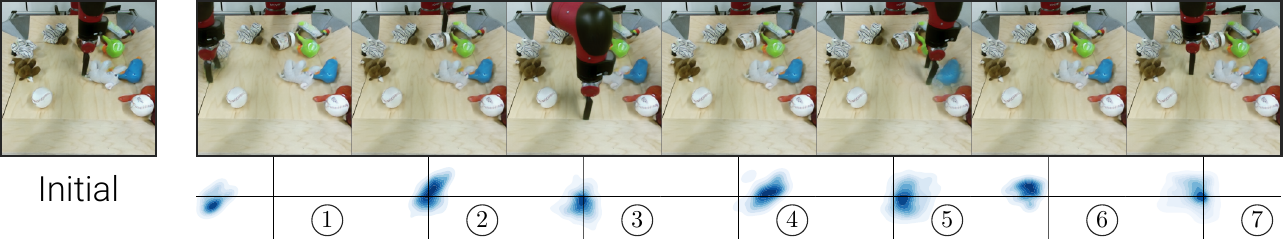}\vspace{-0.2cm}
         \caption{Bair}
     \end{subfigure}\\
      \begin{subfigure}[b]{0.99\textwidth}
         \centering
         \includegraphics[width=\textwidth]{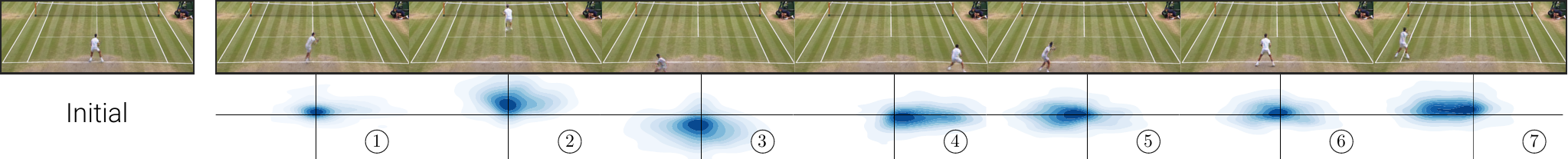}\vspace{-0.2cm}
         \caption{Tennis}
     \end{subfigure}
      \caption{Qualitative evaluation of the action space learned by our model on the three datasets. We consider an initial frame and, for every action, produce a sequence by repeatedly selecting that action and show the final frame.
      The bottom row shows, for each action, the distribution of the displacement $\Delta$ associated with the object of interest.}
    \vspace{-4mm}
    \label{fig:qualitative_actions}
\end{figure*}

\noindent\textbf{Visualization.} Fig.~\ref{fig:qualitative_actions} shows the object motion distribution corresponding to each action learned by CADDY.
On \emph{BAIR}, the model learns actions related to movements on the $x$ (1, 4), $y$ (2, 3) and $z$ (6) axes as well as a no-movement action (7).
On \emph{Atari Breakout}, the model learns actions corresponding to the three possible movements, as well as the inertia of the platform and the physics of the ball and blocks.
Finally, on \emph{Tennis}, the model learns actions corresponding to forward (2) and backward (3) movement, lateral movement (4, 7), no-movement (6) and hitting (1, 5).

\setlength{\belowcaptionskip}{0pt}
\setlength{\belowcaptionskip}{-10pt}

\subsection{Comparison with Previous Works}
\noindent\textbf{Baseline selection.} Since we present the first method for unsupervised PVG, we compare the proposed model against a selection of baselines, adapting existing methods to this new settings. We consider two criteria in the selection of the baselines: the architecture should be adaptable to the new setting without deep modifications, and the source code must be available to include these modifications. Comparing existing video prediction methods in terms of FVD on a 64x64 version of the \emph{BAIR} dataset (quantitative comparison available in \emph{Sup. Mat}) shows that SAVP \cite{lee2018savp} is the best performing method with the code publicly available. Only the methods described in \cite{clark2019dvdgan,luc2020trivdgan,weissenborn2020scaling} are marginally outperforming SAVP but their code is not available.

Furthermore, SAVP \cite{lee2018savp} is a widely used method that features an architecture particularly prone to be adapted to the current setting. During training, SAVP learns an encoder network $E$ that encodes information relative to the transition between successive frames which is used by the generator in the synthesis of the next frame. After training the SAVP model, we cluster the latent space learned by the encoder network on the training sequences using K-means. This procedure produces a set of $K$ centroids that we use as action labels. During evaluation, we compute the latent representations for the input sequence and assign them the index of the nearest centroid.



As a second baseline, we choose MoCoGAN \cite{tulyakov2018moco}. This method possesses favorable characteristics for adaptation to the PVG problem: it separates the content from the motion space and it employs an InfoGAN \cite{chen2016infogan} loss to learn discrete video categories. We consider 6-frame short videos and assume a constant action over them. We employ an action discriminator that predicts the action performed in the input video sequence. This predicted action is used as an auxiliary input to the generator and the InfoGAN loss is used to learn the action space.

Finally, we also include SRVP \cite{franceschi2020stochastic} in our comparison. SRVP is modified similarly to SAVP \cite{lee2018savp} to handle PVG. 
However, due to the poor empirical results and the high training costs, we consider this baseline only on the \emph{BAIR} dataset. Several other works have been considered, but not adopted \cite{kim2019unsupervisedkeypointlearning,xu2018deep} since they would require too important modifications to handle the PVG problem.

SAVP \cite{lee2018savp}, MoCoGAN \cite{tulyakov2018moco} and SRVP \cite{franceschi2020stochastic} are designed for video generation at resolutions lower than our datasets. Therefore, we generate videos at the resolution of 64x80, 64x64 and 128x48 respectively for the \textit{Atari Breakout}, \textit{BAIR} and \textit{Tennis} datasets, and then upscale to full resolution for evaluation. In addition, we create two other baselines, referred to as SAVP+ and MoCoGAN+, obtained by increasing the capacity of the respective networks to generate videos at full resolution. Due to the high memory requirements of SRVP \cite{franceschi2020stochastic}, it was not feasible to produce a version operating at full resolution.

\noindent\textbf{Quantitative evaluation.} We show the results for the evaluation on the \emph{BAIR} dataset in Tab.~\ref{table:bair_results}. According to $\Delta$-\emph{MSE}, MoCoGAN, SAVP and our model learn an action space correlated with the movement of the robot ($\Delta$-\emph{MSE}${<}100$), but our method surpasses these baselines by 26.1\% in $\Delta$-\emph{MSE} and by 24.2\% in $\Delta$-\emph{Acc}, showing greater capacity in learning a set of discrete actions. Finally, SRVP clearly under-performs all the other methods in term of reconstruction (LPIPS), video quality (FID and FVD). Moreover, SRVP predicts a single action class for the whole test set. This degenerated behavior results in a $100\%$  $\Delta$-\emph{ACC} and a poor $\Delta$-\emph{MSE}. Because of these poor results and its very high computational requirement, SRVP is not shown in the following.

Tab.~\ref{table:breakout_results} reports the results on the \emph{Atari Breakout} dataset. Our method outperforms the baselines in both action and video quality metrics. In particular, our model replicates the movement of the platform with an average ADD of 7.29 pixels and an MDR of 2.70\%, indicating that the learned set of actions are consistent with the movement of the user-controlled \emph{Atari Breakout} platform. Note that MoCoGAN obtains a lower MDR but a much higher ADD showing that it generates correctly the platform but at the wrong position.


The evaluation results on the \emph{Tennis} dataset are reported in Tab.~\ref{table:tennis_results}.  With the exception of FVD, where the result is close to SAVP+, our method obtains the best performance in all the metrics. A $\Delta$-\emph{MSE} of 72.2\% shows that actions with a consistent associated movement are learned. On the other hand, the $\Delta$-\emph{MSE} scores of the other baselines show that the movements associated with each action do not present a consistent meaning. Moreover, our method features the lowest ADD and a significantly lower MDR of 1.01\% which indicate that our method consistently generates the player and moves it accurately on the field.

\begin{table}
\begin{center}
\setlength{\tabcolsep}{4pt}
\footnotesize
\begin{tabular}{lccccccc}
\toprule%
 & LPIPS $\downarrow$  & FID $\downarrow$ & FVD $\downarrow$& $\Delta$-\emph{MSE} $\downarrow$ & $\Delta$-\emph{Acc} $\uparrow$\\
\midrule
MoCoGAN \cite{tulyakov2018moco} & 0.466 & 198 & 1380 & 88.8 & 20.7 \\
MoCoGAN+ & 0.201 & 66.1 & 849 & 98.4 & 22.9 \\
SAVP \cite{lee2018savp} & 0.433 & 220 & 1720 & 80.9 & 41.4  \\
SAVP+ & \textbf{0.154}  & \textbf{27.2} & \textbf{303}& 82.0 & 44.8 \\
SRVP \cite{franceschi2020stochastic} & 0.491  & 224 & 3540 & (100) & (100) \\
\midrule
\methodname~(Ours) & 0.202  & 35.9 & 423 & \textbf{54.8} & \textbf{69.0}\\
\bottomrule
\end{tabular}
\end{center}
\caption{Comparison with baselines on the \textit{BAIR} dataset. $\Delta$-\emph{MSE}, $\Delta$-\emph{Acc} and MDR in \%, ADD in pixels. Parentheses indicate degenerated cases resulting in uninformative metrics (see details in the text).}
\label{table:bair_results}
\end{table}

\begin{table}
\begin{center}

\setlength{\tabcolsep}{0.8pt}
\footnotesize
\begin{tabular}{lcccccc}
\toprule
 & LPIPS$\downarrow$  & FID$\downarrow$ & FVD$\downarrow$ & $\Delta$-\emph{MSE}$\downarrow$ & $\Delta$-\emph{Acc}$\uparrow$ & (ADD, MDR)$\downarrow$\\
\midrule
MoCoGAN \cite{tulyakov2018moco} & 0.234 & 99.9 & 447 & 99.6 & 81.9 & (46.0,\textbf{0.795})\\
MoCoGAN+ & 65.8e-3 & 10.4 & 103 & 103 & 57.5 & (54.6,17.4) \\
SAVP \cite{lee2018savp} & 0.239  & 98.4 & 487 & 103 & 58.1 & 24.7, 21.0\\
SAVP+ & 39.3e-3  & 4.84 & 84.4 & 104 & 85.6 & 15.8, 51.5\\
\midrule

\methodname~(Ours) & \textbf{7.66e-3}  & \textbf{0.716} & \textbf{5.94} & \textbf{82.7} & \textbf{91.6} & \textbf{7.29}, 2.70\\

\bottomrule
\end{tabular}
\end{center}
\caption{Comparison with baselines on the \textit{Atari Breakout} dataset. $\Delta$-\emph{MSE}, $\Delta$-\emph{Acc} and MDR in \%, ADD in pixels.}
\label{table:breakout_results}
\end{table}

\begin{table}
\begin{center}

\setlength{\tabcolsep}{1.0pt}
\footnotesize
\begin{tabular}{lcccccc}
\toprule
 & LPIPS$\downarrow$  & FID$\downarrow$ & FVD$\downarrow$& $\Delta$-\emph{MSE}$\downarrow$ & $\Delta$-\emph{Acc}$\uparrow$ & (ADD, MDR)$\downarrow$\\
\midrule
MoCoGAN \cite{tulyakov2018moco} & 0.266 & 132 & 3400 & 101 & 26.4 & 28.5, 20.2 \\
MoCoGAN+ & 0.166 & 56.8 & 1410 & 103 & 28.3 & 48.2, 27.0\\
SAVP \cite{lee2018savp} & 0.245  & 156 & 3270 & 112 & 19.6 & 10.7, 19.7\\
SAVP+ & 0.104 & 25.2 & \textbf{223}& 116 & 33.1 & 13.4, 19.2  \\
\midrule

\methodname~(Ours) & \textbf{0.102}  & \textbf{13.7} & 239 & \textbf{72.2} & \textbf{45.5} & \textbf{8.85}, \textbf{1.01}\\

\bottomrule
\end{tabular}
\end{center}
\caption{Comparison with baselines on the \textit{Tennis} dataset. $\Delta$-\emph{MSE}, $\Delta$-\emph{Acc} and MDR in \%, ADD in pixels.}
\label{table:tennis_results}
\end{table}

\begin{table}
    \centering
    \def\arraystretch{1.0}
    \resizebox{\linewidth}{!}{
    \setlength\tabcolsep{0pt}
    \footnotesize
    \renewcommand{\arraystretch}{0.5}
    \hspace{-5mm}
    \resizebox{0.607\linewidth}{!}{
    \begin{tabular}{l@{\hskip 1mm}ccc}
         & $t=1$ & $t=5$ & $t=10$  \\
         \rotatebox{90}{\hspace{3.0mm}\scalebox{1.1}{Original}} & 
         \includegraphics[trim=64 128 64 2,clip,width=0.2\columnwidth]{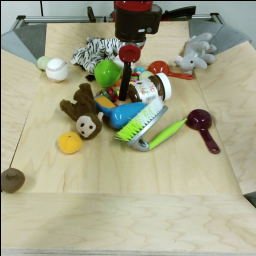} & 
         \includegraphics[trim=64 128 64 2,clip,width=0.2\columnwidth]{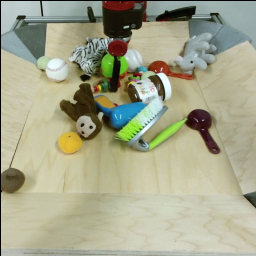} & 
         \includegraphics[trim=64 128 64 2,clip,width=0.2\columnwidth]{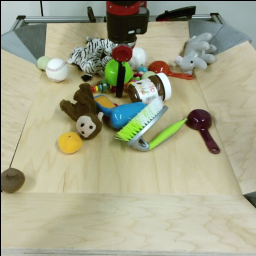} \\
         
         \rotatebox{90}{\hspace{0mm}\scalebox{1.1}{MoCoGAN+}} & \includegraphics[trim=64 128 64 2,clip,width=0.2\columnwidth]{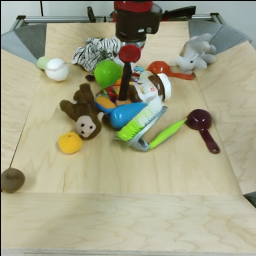} & 
         \includegraphics[trim=64 128 64 2,clip,width=0.2\columnwidth]{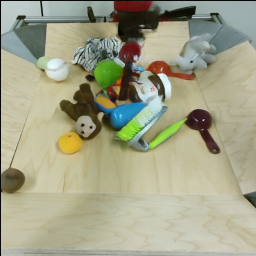} & 
         \includegraphics[trim=64 128 64 2,clip,width=0.2\columnwidth]{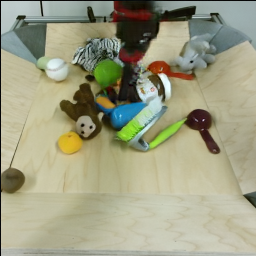} \\
         
         \rotatebox{90}{\hspace{3.5mm}\scalebox{1.1}{SAVP+}} & 
         \includegraphics[trim=64 128 64 2,clip,width=0.2\columnwidth]{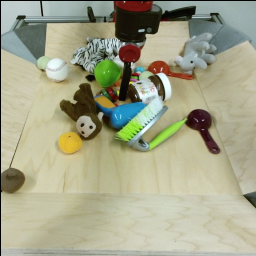} & 
         \includegraphics[trim=64 128 64 2,clip,width=0.2\columnwidth]{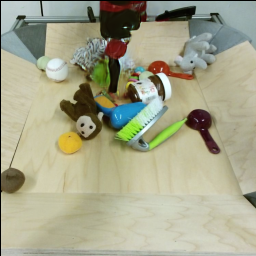} & 
         \includegraphics[trim=64 128 64 2,clip,width=0.2\columnwidth]{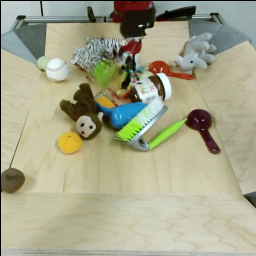} \\
         
         \rotatebox{90}{\hspace{5mm}\scalebox{1.1}{Ours}} & 
         \includegraphics[trim=64 128 64 2,clip,width=0.2\columnwidth]{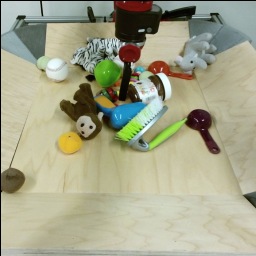} & 
         \includegraphics[trim=64 128 64 2,clip,width=0.2\columnwidth]{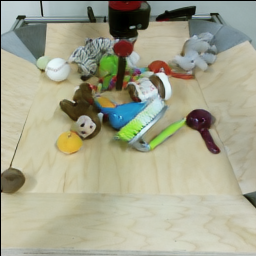} & 
         \includegraphics[trim=64 128 64 2,clip,width=0.2\columnwidth]{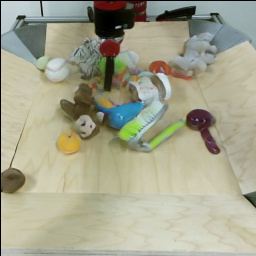} \\

    \end{tabular}
    }
    \quad
    \begin{tabular}{l@{\hskip 0.7mm}ccc}
         & $t=1$ & $t=8$ & $t=16$  \\
         \rotatebox{90}{\hspace{2.5mm}Original} & \includegraphics[trim=0 0 150 0,clip,width=0.2\columnwidth]{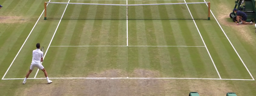} &
         \includegraphics[trim=0 0 150 0,clip,width=0.2\columnwidth]{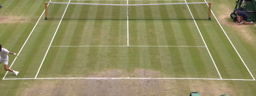} & 
         \includegraphics[trim=0 0 150 0,clip,width=0.2\columnwidth]{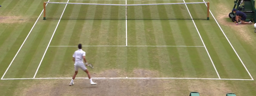}  \\
         
         \rotatebox{90}{\hspace{0mm}MoCoGAN+} & \includegraphics[trim=0 0 150 0,clip,width=0.2\columnwidth]{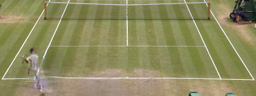} & 
         \includegraphics[trim=0 0 150 0,clip,width=0.2\columnwidth]{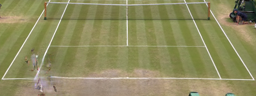} & 
         \includegraphics[trim=0 0 150 0,clip,width=0.2\columnwidth]{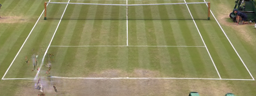}  \\
         
         \rotatebox{90}{\hspace{3.5mm}SAVP+} & 
         \includegraphics[trim=0 0 150 0,clip,width=0.2\columnwidth]{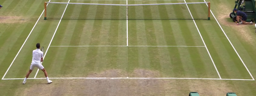} & 
         \includegraphics[trim=0 0 150 0,clip,width=0.2\columnwidth]{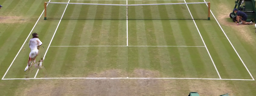} & 
         \includegraphics[trim=0 0 150 0,clip,width=0.2\columnwidth]{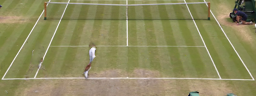}  \\
         
         \rotatebox{90}{\hspace{5mm}Ours} & 
         \includegraphics[trim=0 0 150 0,clip,width=0.2\columnwidth]{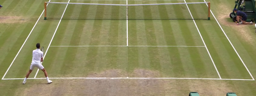} & 
         \includegraphics[trim=0 0 150 0,clip,width=0.2\columnwidth]{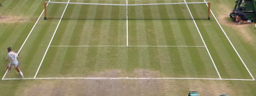} & 
         \includegraphics[trim=0 0 150 0,clip,width=0.2\columnwidth]{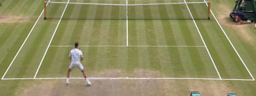}  \\

    \end{tabular}
    }
    \captionof{figure}{Reconstruction results on the \emph{BAIR} (left) and \emph{Tennis} (right) datasets. We zoom in for better visualization.}
    \vspace{-3mm}
    \label{fig:qualitative_bair_tennis_reconstruction}
\end{table}

\noindent\textbf{Qualitative evaluation} In Fig.~\ref{fig:qualitative_bair_tennis_reconstruction}, we show examples of reconstructed sequences on the \emph{BAIR} and \emph{Tennis} datasets. Our method achieves a precise placement of the object of interest with respect to the input sequence. In the \emph{Sup. Mat.} we show additional qualitative results as well as sample videos generated interactively.



\noindent\textbf{User evaluation.} To complete our evaluation, we perform a user study on the \emph{Tennis} dataset. We sample 23 random frames that we use as initial frames. For each of them, we generate $K$ continuations of the sequences, one for each action, each produced by repeatedly selecting the corresponding action. For each sequence, users are asked to select the performed action among a predefined set (\emph{Left},~\emph{Right},~\emph{Forward},~\emph{Backward},~\emph{Hit the ball}~or \emph{Stay}). An additional option \emph{Other} is provided in case the user cannot recognize any action. We measure user agreement using the Fleiss' kappa measure \cite{fleiss1971measuring} that is commonly used to evaluate agreement between categorical ratings~\cite{fleiss1981measurement}. 
In addition, to validate that all the actions can be generated (\ie detecting mode collapse), we compute the diversity of the action space, expressed as the entropy of the user-selected actions. Results are shown in Tab.~\ref{table:user_study}. While all methods capture actions with high diversity, our method shows a higher agreement, indicating that users consistently associate the same action label to each learned action. Looking at the details of the votes for our method (see \emph{Supp. Mat.}), we observe that most disagreements are due to cases where the player hits the ball while moving. In contrast, the other methods do not obtain high user agreement, with actions assuming different meanings, depending on the particular frame used as context. 

\begin{table}
\begin{center}

\footnotesize
\begin{tabular}{lccccc}
\toprule
 & Agreement & Diversity & \emph{Other} votes\\
\midrule
MoCoGAN \cite{tulyakov2018moco} & -3.15e-3 & 1.58 & 1.80\% \\
MoCoGAN+ & -2.84e-3 & 1.51 & 28.0\% \\
SAVP \cite{lee2018savp} & 0.0718 & 1.69 & 7.14\% \\
SAVP+ & -1.97e-3 & \textbf{1.80} & 5.40\% \\
\midrule

\methodname~(Ours) & \textbf{0.469} & 1.65 & 1.61\% \\

\bottomrule
\end{tabular}
\end{center}
\caption{User study results on the \textit{Tennis} dataset.}
\vspace{-3mm}
\label{table:user_study}
\end{table}


\vspace{-3mm}
\section{Conclusions}
In this work, we propose the unsupervised learning problem of playable video generation. We introduce CADDY, a self-supervised method based on an encoder-decoder architecture that uses predicted action labels as a bottleneck. We evaluate our method on three varied datasets and show state-of-the-art performance. Our experiments show that we can learn a rich set of actions that offer the user a gaming-like experience to control the generated video. As future work, we plan to extend our method to multi-agent environments.

{\small
\bibliographystyle{ieee_fullname}
\bibliography{egbib}
}

\end{document}